\documentclass[letterpaper, 10 pt, conference]{ieeeconf}  

\IEEEoverridecommandlockouts                              
\usepackage[utf8]{inputenc}
\usepackage[T1]{fontenc}

\usepackage{cite}
\usepackage[pdftex]{graphicx}
\usepackage{amsmath}
\usepackage{amssymb}
\usepackage{array}
\usepackage{wasysym}
\usepackage{bm}
\usepackage{siunitx}
\usepackage{url}
\usepackage{color}
\usepackage{epstopdf}
\usepackage[ruled]{algorithm2e} 
\usepackage{arydshln}
\usepackage{mathtools}

\usepackage{stfloats}
\usepackage{relsize}
\usepackage{cancel}
\usepackage{epstopdf}
\usepackage{environ}
\usepackage{amsthm}
\usepackage{cite}
\usepackage{hyperref}
\usepackage{balance}

\NewEnviron{ruledtable}{%
\par\addvspace{2mm}\hrule height 0.03cm 
\begin{table}[!h]\BODY\end{table}
\hrule height 0.03cm \addvspace{2mm}
}

\DeclareMathOperator*{\argmin}{arg\,min}

\theoremstyle{plain}

\newtheorem{definition}{Definition}

\usepackage{CJKutf8}

\begin{document}
\title{\LARGE \bf{Safety-critical Locomotion of Biped Robots in Infeasible Paths:
Overcoming Obstacles during Navigation toward Destination}}
\author{Jaemin Lee$^{1}$, Min Dai$^{2}$, Jeeseop Kim$^{2}$, and Aaron D. Ames$^{2}$
\thanks{$^{1}$J. Lee is with the Department of Mechanical and Aerospace Engineering, North Carolina State University, Raleigh, NC, USA} 
\thanks{$^{2}$M. Dai, J. Kim, and A.D. Ames are with the Department of Mechanical and Civil Engineering, California Institute of Technology, Pasadena, CA, USA} 
}

\maketitle

\begin{abstract}
This paper proposes a safety-critical locomotion control framework employed for legged robots exploring through infeasible path in obstacle-rich environments. Our research focus is on achieving safe and robust locomotion where robots confront unavoidable obstacles en route to their designated destination. Through the utilization of outcomes from physical interactions with unknown objects, we establish a hierarchy among the safety-critical conditions avoiding the obstacles. This hierarchy enables the generation of a safe reference trajectory that adeptly mitigates conflicts among safety conditions and reduce the risk while controlling the robot toward its destination without additional motion planning methods. In addition, robust bipedal locomotion is achieved by utilizing the Hybrid Linear Inverted Pendulum model, coupled with a disturbance observer addressing a disturbance from the physical interaction. 
\end{abstract}

\section{Introduction}
\label{section1}
\begin{figure}[t] 
\centering
\includegraphics[width=0.95\linewidth]{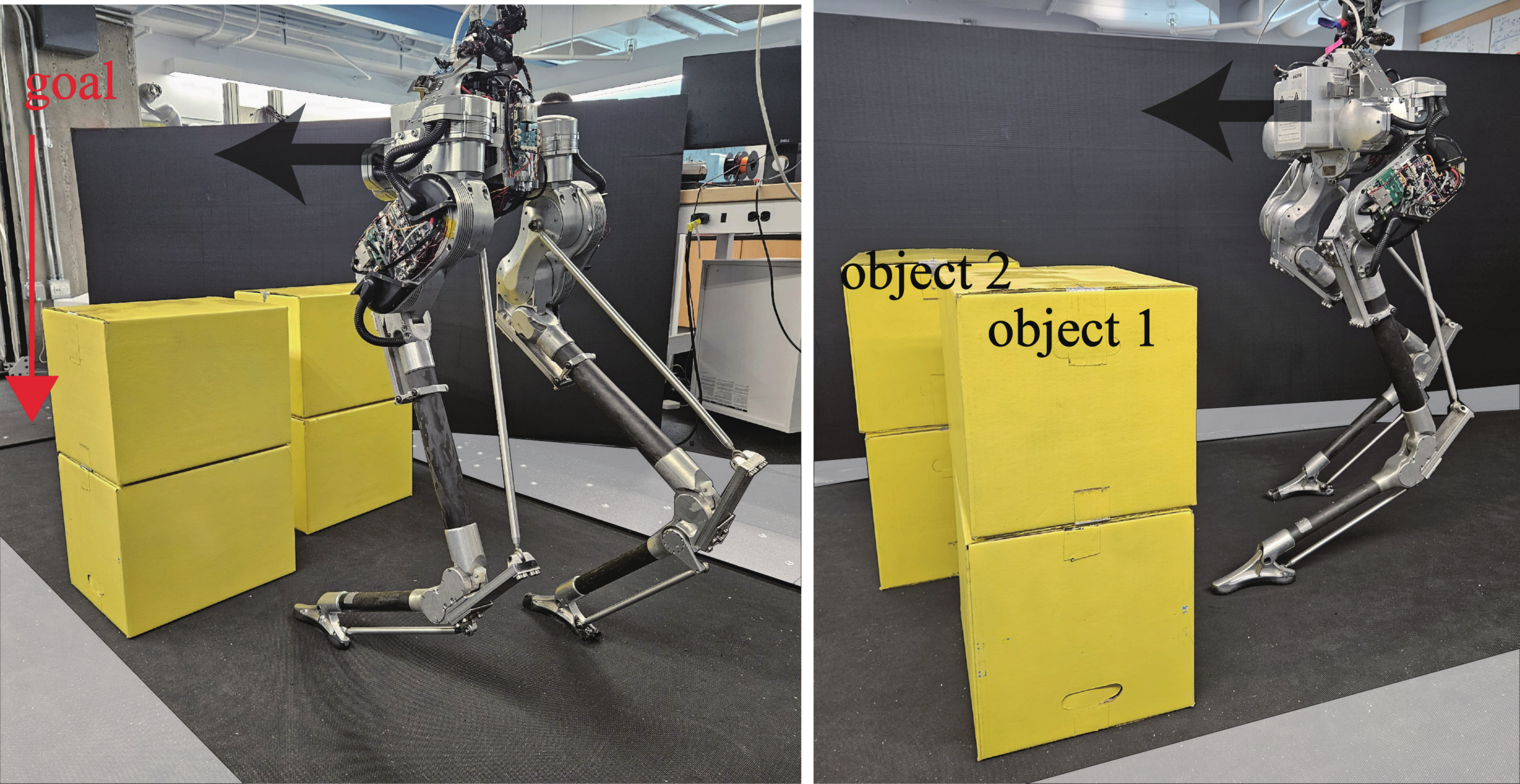}
\vspace{-2mm}
\caption{\textbf{Safety-critical locomotion scenario}: Our framework enables safe locomotion when the robot is blocked by unknown objects.}
\label{Fig0}
\vspace{-0.7cm}
\end{figure}

Legged robots have been extensively employed in navigating challenging terrains and obstacle-rich environments. Their reliability as platforms becomes increasingly evident in complex and crowded settings. In such demanding scenarios, the significance for safety-critical locomotion emerges as an important requirement to ensure the successful completion of given missions. Our paper tackles a unique challenge—navigating toward a goal destination without any existing safe and feasible path, necessitating the negotiation of obstacles (depicted in Fig. \ref{Fig0}). This study introduces a novel approach that achieves the goal destination through physical interactions with obstacles, simultaneously enhancing safety, by using hierarchical safety-critical control and robust walking patterns based on the hybrid linear inverted pendulum (H-LIP) \cite{xiong20223} model and disturbance observer (DOB).

\subsection{Related Works}
The locomotion control paradigm commonly involves multiple layers, including high-level planning, walking pattern design, whole-body control, low-level joint space control, and so on. Each layer employs either reduced-order or full-order models utilized to its specific objectives and requirements. In general, reduced-order models (RoMs) are used to capture the dynamic behaviors in high-level planning and walking pattern generation. For instance, the widely utilized linear inverted pendulum (LIP) model features prominently in walking pattern generation processes like time-to-velocity reversal planning \cite{kim2020dynamic}, Step-to-Step dynamics-based planning \cite{xiong20223}, and capture point dynamics-based planning \cite{englsberger2011bipedal}. For these methods, we frequently assume that nominal safe and tractable trajectories of center of mass (CoM) can be predefined then focus on generating swing foot trajectories preventing the robots from falling. Once the CoM and swing foot trajectories are designed, whole-body controller (WBCs) \cite{lee2012intermediate,kuindersma2016optimization,righetti2013optimal} are employed to generate dynamically consistent control commands to track these trajectories, considering contacts, full-body dynamics, and constraints.

In the pursuit of safe legged locomotion, extensive research has focused on high-level motion planning using linear temporal logic \cite{warnke2020towards}, model predictive control \cite{gaertner2021collision, chiu2022collision, narkhede2022sequential, naveau2016reactive}, learning techniques \cite{csomay2021episodic, ji2022concurrent, li2021reinforcement}, and safety-critical planning (control) \cite{liao2022walking, grandia2021multi, kim2023safety, lee2024safety_ICRA, lee2024safety, kim2024safety_ICRA}. While the above methods show efficient locomotion when there exists a safe path toward the goal destination, it is a significant challenge when there is no safe path from the initial position to the goal due to the multiple safety conditions or constraints \cite{escande2014hierarchical, lindemann2019control}. Our previous work in \cite{lee2023hierarchical} introduces a hierarchy among multiple safety critical conditions, allowing violation of less important conditions to resolve contradictions while strictly satisfying more crucial ones, which logically makes sense in complex environments.

Properly considering interactions with environments or other agents is also crucial for safe locomotion. Two main key issues for the safe locomotion are 1) estimating unknown properties of objects and 2) making the systems robust by compensating for disturbance. Firstly, the estimation of unknown objects often employs rigid-body dynamics for dexterous manipulation of robots \cite{yu2005estimation, fazeli2017parameter, doshi2022manipulation}, which can be adapted to legged robots with precise control of contact forces. The legged robots less dexterously handle the objects thus we need to simplify the estimation problem by addressing several assumptions. For instance, we need to assume that geometry information of the object is given and the density is even then we can reduce the number of variables of the object's inertial property \cite{mavrakis2020estimation}. 

Secondly, interactive forces act as disturbances to the robot so that feedback gains are dynamically adjusted to stabilize the robot in real-time for robust full-body feedback controllers \cite{lee2022online, djeha2023robust,dai2023data}. In addition, the application of DOB contributes to the robustness of legged robots, ensuring balance maintenance \cite{sato2008zmp} and stability during locomotion \cite{czarnetzki2009observer}, drawing insights from zero moment point movements. Moreover, H-LIP model is beneficial to consider the system as two linear systems in frontal and sagittal planes, respectively, so it is intuitive and straightforward to apply DOB in the walking pattern generation using H-LIP model.  

\subsection{Contributions}

\begin{figure*}[t] 
\centering
\includegraphics[width=0.95\linewidth]{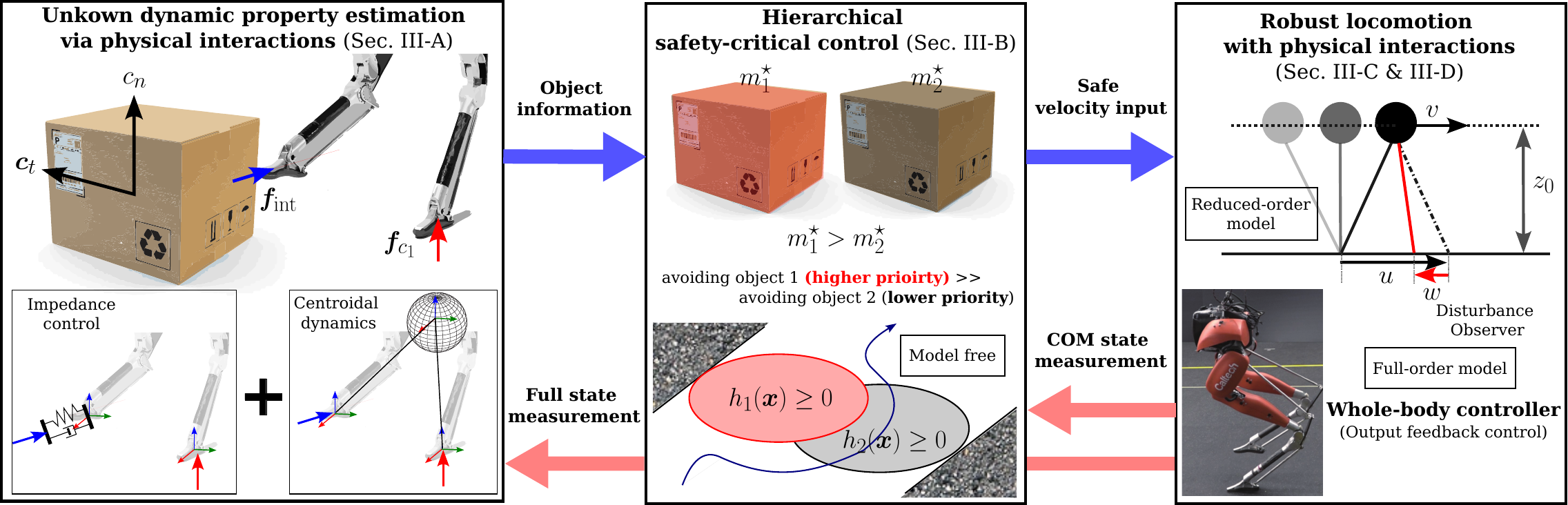}
\caption{\textbf{Structure of the proposed framework}: The first layer focuses on estimating the unknown dynamic properties of objects (Section \ref{section3_1}). With the estimated properties, a hierarchy among the safety conditions is established, and the safe velocity input is computed in the second layer (Section \ref{section3_2}). The last layer is dedicated to robust locomotion involving the use of H-LIP with DOB (Section \ref{section3_3}) and the WBC (Section \ref{Section3_4}).}
\label{Fig1}
\vspace{-0.55cm}
\end{figure*}

The proposed approach consists of three layers, as illustrated in Fig. \ref{Fig1}. The dynamic properties of objects are estimated approximately through physical interactions with them. a priority among safety conditions is defined for avoiding heavier objects over lighter ones. In this framework, we design a safety-critical planning layer using a double-integrator system model and CBFs, following the method in \cite{lee2023data}. The middle layer features a walking pattern generator that follows desired CoM trajectories using the H-LIP model and DOB \cite{shim2021disturbance}, compensating for any misalignment of the swing foot caused by these interactions. Based on the generated walking pattern, the WBC computes torque commands to track the task trajectories.

Our contributions are threefold. First, we introduce a multi-layered framework that enables legged robots to achieve robust and safe locomotion by leveraging physical interactions between the robot's foot and objects when facing infeasible paths. In the proposed framework, additional motion planning methods are not required for the safe locomotion. Second, the robot can autonomously differentiate between heavier and lighter objects and prioritize avoiding the heavier one, ensuring safer locomotion when no clear path is available while reaching the goal. This is done by mitigating contradictions in the safety conditions when the safety-critical controller is utilized. Finally, the use of H-LIP with DOB compensates for disturbances caused by physical interactions, allowing the robot to generate more robust locomotion.

The structure of the paper is as follows: Section \ref{section2} outlines the theoretical background and preliminaries. In Section \ref{section3}, we specify the four subsections of the proposed layered architecture: 1) physical interaction and estimation, 2) safety-critical, model-free planning, 3) H-LIP with DOB, and 4) output feedback control (WBC). Section \ref{section4} presents high-fidelity simulations demonstrating the effectiveness of our methods using a bipedal robot ``Cassie''.

\section{Preliminary}
\label{section2}
This section briefly recalls the fundamental concepts of CBFs and the H-LIP model for bipedal walking.  

\subsection{Control Barrier Function}
This section reviews the fundamental concepts of CBFs. We begin by defining a safe set $\mathcal{C} \subseteq \mathbb{R}^{n}$, which is the $0$-superlevel set of a continuously differentiable function $h: \mathbb{R}^{n} \to \mathbb{R}$ such that $\mathcal{C} \coloneqq \{\bm{x}\in \mathbb{R}^{n}: h(\bm{x}) \geq 0 \} $. In addition, we consider a state space $\mathcal{X} \subseteq \mathbb{R}^{n}$ and an input space $\mathcal{U} \subseteq \mathbb{R}^{m}$. With thees definitions, a nonlinear control affine system is expressed as follows:
\begin{equation}
    \dot{\bm{x}} = f(\bm{x}) + g(\bm{x}) \bm{u}
\end{equation}
where $\bm{x} \in \mathcal{X}$, $\bm{u} \in \mathcal{U}$, $f: \mathcal{X} \to \mathbb{R}^{n}$, and $g: \mathcal{X} \to \mathbb{R}^{n\times m}$. It is noted that $f$ and $g$ are locally Lipschitz continuous. According to \cite{ames2016control}, we define a CBF as follows:
\begin{definition}
    Given a safe set $\mathcal{C} = \{\bm{x} \in \mathcal{X}: h(\bm{x})\geq 0 \}$, $h$ is a control barrier function (CBF) if there exists an extended class $\mathcal{K}$ function $\alpha\in \mathcal{K}^{e}_{\infty}$ such that for all $\bm{x} \in  \mathcal{C}$:
    \begin{equation*}
        \sup_{\bm{u}\in \mathcal{U}} \dot{h}(\bm{x}, \bm{u}) = \sup_{\bm{u} \in \mathcal{U}} \left[\mathcal{L}_f h(\bm{x}) + \mathcal{L}_{g} h(\bm{x}) \bm{u}) \right] \geq - \alpha(h(\bm{x}))
    \end{equation*}
    where $\mathcal{L}_{f} h(\bm{x}) = \frac{\partial h}{\partial \bm{x}}(\bm{x}) f(\bm{x})$ and $\mathcal{L}_g h(\bm{x}) = \frac{\partial h}{\partial \bm{x}}(\bm{x}) g(\bm{x})$ are Lie derivatives. 
\end{definition}

The above definition offers a simple and concise formulation of the safety filter, which is in a form of the quadratic program, when the desired control input $\bm{u}^{d}$ is computed a priori. However, it is significantly important to formalize the CBF $h$ and the corresponding function $\alpha$ based on the state constraints. In this paper, we design and validate CBFs using a data-driven method, as proposed in \cite{lee2023data}.

\subsection{Hybrid Linear Inverted Pendulum}
As proposed in \cite{xiong20223}, bipedal locomotion can be implemented by using the H-LIP model, which consists of a single support phase (SSP) and a double support phase (DSP) in a predefined sequence. We can describe the dynamic equations of H-LIP for each walking phase as follows:
\begin{equation}
    \begin{split}
        \textrm{SSP:} &\quad \ddot{p} = \lambda^{2} p, \quad \textrm{DSP:} \quad \ddot{p} =0
    \end{split}
\end{equation}
where $p$ is the mass position relative to the stance foot location and $\lambda = \sqrt{\frac{g}{z_0}}$ with the height of mass $z_0$. The state-space model for the SSP is described as follows:
\begin{equation}
    \underbrace{\frac{d}{dt}\left[\begin{array}{c} p \\ v \end{array} \right]}_{\dot{\mathbf{x}}_{\textrm{SSP}}} = \underbrace{\left[\begin{array}{cc} 0 & 1 \\ \lambda^{2} &0 \end{array} \right]}_{\mathbf{A}_{\textrm{SSP}}} \underbrace{\left[\begin{array}{c} p \\ v \end{array} \right]}_{\mathbf{x}_{\textrm{SSP}}},
\end{equation}
where $v$ denotes the velocity of mass. After the single support phase duration $T_{\textrm{SSP}}$, the initial state for the next step is obtained analytically: 
\begin{equation}
    \mathbf{x}_{\textrm{SSP}}^{-} = e^{\mathbf{A}_{\textrm{SSP}} T_{\textrm{SSP}}} \mathbf{x}_{\textrm{SSP}}^{+},
\end{equation}
where $+$ and $-$ denote the states after and before the transition, respectively. The state at the $k+1$-th step is written in terms of the $k$-th state and the step size $u_{[k]}$ as follows:
\begin{gather}
    \mathbf{x}_{\textrm{SSP}[k+1]}^{H} = \mathbf{A}^{H} \textbf{x}_{\textrm{SSP}[k]}^{H} + \mathbf{B}^{H} u_{[k]}^{H} \\
    \mathbf{A}^{H} = e^{\mathbf{A}_{\textrm{SSP}} T_{\textrm{SSP}}}\left[\begin{array}{cc} 1 & T_{\textrm{DSP}} \\ 0 & 1 \end{array} \right], \quad  \mathbf{B}^{H} = e^{\mathbf{A}_{\textrm{SSP}} T_{\textrm{SSP}}}\left[\begin{array}{c} -1 \\ 0 \end{array} \right], \nonumber
\end{gather}
where the superscript $(.)^{H}$ represents properties associated with the H-LIP model. When $T_{\mathrm{DSP}}=0$, it is noted that the H-LIP model becomes identical to the standard LIP model consisting of only one domain.

Given the robot's CoM state $\mathbf{x}^{R} = [p^{R}, \; v^{R}]^{\top}$, where the CoM position and velocity computed by the joint measurements, we introduce the step-to-step dynamics of the robot: $\mathbf{x}_{[k+1]}^{R} = \mathcal{P}_x(\bm{q}_{[k]}^{-}, \dot{\bm{q}}_{[k]}^{-}, \bm{\tau}(t))$ where $\bm{q} \in \mathcal{Q} \subset \mathbb{R}^{n_q}$, $\dot{\bm{q}} \in \mathbb{R}^{n_q}$, and $\bm{\tau}(t)\in \Gamma \subset \mathbb{R}^{n_q-6}$ denote joint position, velocity, and torque command, respectively. Then the dynamic model is approximated with the model discrepancy between the H-LIP and full-body models as follows:
\begin{equation} \label{H-LIP}
    \mathbf{x}_{[k+1]}^{R} = \mathbf{A}^{H} \mathbf{x}_{[k]}^{R} + \mathbf{B}^{H} u_{[k]}^{R} + w_{[k]}
\end{equation}
where $w_{[k]} = \mathcal{P}_{x}(\bm{q}_{[k]}^{-}, \dot{\bm{q}}_{[k]}^{-}, \bm{\tau}(t)) - \mathbf{A}^{H} \mathbf{x}_{[k]}^{R} - \mathbf{B}^{H} u_{[k]}^{R}$. As noted in \cite{xiong20223}, when the discrepancy $w_{k}$ is bounded, the robot's stepping input is computed by the following controller:  
\begin{equation} \label{LIP-controller}
    u_{[k]}^{R} = u_{[k]}^{H} + \mathbf{K}_{\textrm{step}} ( \mathbf{x}_{[k]}^{R} - \mathbf{x}_{[k]}^{H})
\end{equation}
where $\mathbf{K}_{\textrm{step}}$ is the gain matrix such that $\mathbf{A}^{H} + \mathbf{B}^{H} \mathbf{K}_{\textrm{step}}$ is Hurwitz. In this paper, we compute $\mathbf{K}_{\textrm{step}}$ as linear quadratic regulator (LQR) gain by solving the Riccati equation. The detailed statement for the disturbance $w_k$ (model discrepancy) is addressed by referring to the forward invariant and bounded set \cite{xiong20223}. In this study, we assume that $\mathbf{K}_{\textrm{step}}$ and the step size are properly selected to make the discrepancy $w_k$ negligible. Thus, we assume that the only source of disturbance is the physical interaction between the robot and unknown objects.


\section{Multi-layered Architecture}
\label{section3}
This section describes the detailed architecture of the proposed approach. First, we explain how to estimate the dynamic properties of unknown objects using interactive force, without relying on force/torque sensors. Second, based on the estimated object parameters, we establish a hierarchy of safety conditions and compute the reference velocity using a safety filter accordingly. Third, the H-LIP model generates the walking pattern for the legged robot, incorporating DOB. Lastly, we design the low-level controller (WBC) for tracking the task trajectory for safe walking of the robot. 

\subsection{Physical Interaction and Estimation}
\label{section3_1}
In this paper, we assume that objects, which are rigid bodies with unknown mass ($m$) and inertia matrix ($\mathcal{I}$), are rested on the plain ground with surface contacts. Then, given a twist $\bm{\mathcal{V}}_{b} = (\bm{\omega}_{b},\bm{v}_{b} )$ and a wrench $\bm{\mathcal{F}}_{b} = (\bm{\mathcal{M}}_{b}, \bm{f}_{b})$ with respect to the body frame (CoM frame), the dynamic equation of a single rigid body (SRB) is expressed as follows:
\begin{gather} \label{srb_dynamics}
    \bm{\mathcal{F}}_{b} = \mathcal{G}_{b} \dot{\bm{\mathcal{V}}}_{b} - [\textrm{ad}_{\bm{\mathcal{V}}_{b}}]^{\top} \mathcal{G}_{b} \bm{\mathcal{V}}_{b} \\
    \mathcal{G}_{b} = \left[\begin{array}{cc} \mathcal{I}_{b} & \mathbf{0}_{3\times 3} \\ \mathbf{0}_{3\times 3} & m \mathbf{I}_{3\times 3} \end{array} \right], \quad \left[ \textrm{ad}_{\bm{\mathcal{V}}_b} \right] = \left[\begin{array}{cc} \left[ \bm{\omega}_{b} \right]_{\times} & \mathbf{0}_{3\times 3} \\ \left[ \bm{v}_{b} \right]_{\times} & \left[\bm{\omega}_{b} \right]_{\times} \end{array} \right] \nonumber
\end{gather}
where $\left[ a \right]_{\times}$ denotes the skew symmetric matrix of $a\in \mathbb{R}^{3}$. We make three assumptions to simplify the process:
\begin{itemize}
    \item The geometry of the object is known, and the mass is uniformly distributed, though its value is unknown.
    \item There is a single interaction force between the robot and the object at a solid contact point.
    \item The friction coefficient between the ground and the object is known.
\end{itemize}

When a force/torque sensor is unavailable at the contact point, we need to approximately estimate the interaction force $\hat{\bm{f}}_{\textrm{int}}$ and transform it into the wrench $\bm{\mathcal{F}}_{b}$. Our idea is to estimate the interaction force using centroidal dynamics, which is similar to the method in \cite{lee2022online}, assuming that only the swing foot contacts the object while the stance foot remains on the ground. The centroidal dynamics is written by:
\begin{gather} \label{centroidal}
    \bm{f}_{\textrm{int}} + \bm{f}_{\textrm{ground}} = M^{R} (\ddot{\bm{p}} - \bm{g}),\\
     \dot{\mathbf{L}} = (\bm{p}_{\textrm{ground}} - \bm{p}) \times \bm{f}_{\textrm{ground}} + (\bm{p}_{\textrm{int}} - \bm{p}) \times \bm{f}_{\textrm{int}} 
\end{gather}
where $M^{R}$, $\mathbf{L}$, $\bm{g}$, and $\bm{f}_{\textrm{ground}}$ denote the total mass of the robot, angular momentum around the CoM, the gravity vector, and the ground contact force, respectively. The positions $\bm{p}$, $\bm{p}_{\textrm{ground}}$, and $\bm{p}_{\textrm{int}}$ refer to the CoM, the stance foot, and the contact point with the object. The estimated interaction force ($\bm{f}_{\textrm{int}}^{\textrm{cm}}$) using the above centroidal dynamics is approximately computed using the nominal ground force ($\bm{f}_{\textrm{ground}}$) in our whole-body controller.

Since the estimated interaction force $\bm{f}_{\textrm{int}}^{\textrm{cm}}$ may not fully account for the internal forces or inaccuracies in $\bm{f}_{\textrm{ground}}$, we need to refine it using task-space controllers, which are formalized using the impedance control as follows:
\begin{gather}
    \mathbf{K}_{p, \textrm{int}}\bm{e}_{\textrm{int}} + \mathbf{K}_{d,\textrm{int}} \dot{\bm{e}}_{\textrm{int}} + \bm{\Lambda}_{\textrm{int}}(\bm{q}) \ddot{\bm{e}}_{\textrm{int}} = \bm{f}_{\textrm{int}} 
\end{gather}
where $\bm{\Lambda}_{\textrm{int}}$, $\mathbf{K}_{p,\textrm{int}}$ and $\mathbf{K}_{d,\textrm{int}}$ denote the mass/inertia matrix, the stiffness and damping coefficients in the task space. The position error $\bm{e}{\textrm{int}} = \bm{x}_{\textrm{sw}}^d - \bm{x}_{\textrm{sw}}$ is calculated using the desired and measured positions of the swing leg, $\bm{x}_{\textrm{sw}}^d$ and $\bm{x}_{\textrm{sw}}$. The task-space impedance control produces $\bm{f}_{\textrm{int}}^{\textrm{imp}}$. Then, we combine the estimated interaction forces by using both centroidal dynamics and task-space impedance control:
\begin{equation}
    \hat{\bm{f}}_{\textrm{int}} = \gamma \bm{f}_{\textrm{int}}^{\textrm{cm}} + (1- \gamma)\bm{f}_{\textrm{int}}^{\textrm{imp}}
\end{equation}
where $0 \leq \gamma \leq 1$ is a weighting factor, heuristically chosen to balance the estimates. The estimation with this weighting offers better tendency compared with one using the centroidal dynamics model.

To estimate the object's mass, we simplify the SRB dynamics by focusing on the mass as the only unknown parameter and considering the linear part of the SRB dynamics based on the assumptions. After we gather non-zero $N_{\textrm{est}}$ data, then, we formulate a least-square error minimization problem to obtain the variable $m$: 
\begin{equation} \label{estimation}
    \begin{split}
        m^{\star(i)} = \argmin_{m} & \quad \|\hat{\bm{f}}_{b,xy}^{(i)} - m (\dot{\bm{v}}_{b,xy}^{(i)} +  \mu g \overline{\bm{v}}_{b,xy}^{(i)}) \|^{2} \\
        \textrm{s.t.} & \quad m \geq 0.
    \end{split}
\end{equation}
where the subscript $(.)_{xy}$ denotes the vector including the $x$ and $y$ components of $(.)$. In addition. $\overline{(.)}$ represents the normalized vector of $(.)$. The final estimated mass is averaged across all samples: $m^{\star} = \frac{\sum_{i=1}^{N_{\textrm{est}}} m^{\star(i)}}{N_{\textrm{est}}}$. This estimation approach, while simple and intuitive, may suffer from overfitting and parameter tuning challenges during impact contacts. Therefore, we focus on relative comparisons of mass properties between objects rather than precise identification.   

\subsection{Safety-critical Model-Free Planning}
\label{section3_2}
The safety-critical planning layer employs a double-integrator system to capture the kinematic characteristics of the robotic system without accounting for the detailed dynamic properties. Using the horizontal base position $\bm{\varphi}\in \mathbb{R}^{2}$, the equation of motion for the double-integrator system is expressed with the state vector $\bm{x}= [\bm{\varphi}^{\top}, \dot{\bm{\varphi}}^{\top}]^{\top} $ as follows:
\begin{equation}
    \underbrace{\left[\begin{array}{c} \dot{\bm{\varphi}} \\ \ddot{\bm{\varphi}} \end{array} \right]}_{\dot{\bm{x}}} = \underbrace{\left[\begin{array}{cc} \mathbf{0} & \mathbf{I} \\ \mathbf{0} & \mathbf{0} \end{array} \right]\left[\begin{array}{c} \bm{\varphi} \\ \dot{\bm{\varphi}} \end{array} \right]}_{f(\bm{x})} + \underbrace{\left[ \begin{array}{c} \mathbf{0} \\ \mathbf{I} \end{array} \right]}_{g(\bm{x})} \bm{u}
\end{equation}
Given a desired trajectory of CoM, $\bm{\varphi}^{d}$, the desired control input is computed as $\bm{u}^{d} =k(\bm{\varphi}^{d}, \bm{\varphi})$ where $k: \mathbb{R}^{2} \times \mathbb{R}^{2} \to \mathbb{R}^{2}$ is Lipschitz continuous. To ensure safety, we formalize multiple CBFs in terms of the position of base and the $i$-th object position $\bm{\mathcal{B}}_{i} \in \mathbb{R}^{2}$. The CBFs are designed by referring the point of each object and the geometric information. 
\begin{equation}
    h_{i}(\bm{x}) = - \|\bm{\varphi} - \bm{\mathcal{B}}_{i} \|^{2} + (r_{i}^{2} + \zeta_{i}) \geq 0
\end{equation}
where $r_{i}\geq 0$ denotes the maximum distance between the center and the vertex of the $i$-th object on the ground plane. The term $\zeta_i \geq 0$ is a conservative value derived using the data-driven method from \cite{lee2023data}, given the bounded velocity range $[\dot{\bm{\varphi}}_{\min}, \dot{\bm{\varphi}}_{\max}]$.

We consider two assumptions in the problem statements: 1) The safety conditions are contradictory on the path while achieving the goal, and 2) A hierarchy exists between them ($h_1 \gg h_2$), based on the mass property estimated from the interaction force in \eqref{estimation}. More specifically, avoiding the heavier object (related to $h_1$) takes precedence over avoiding the lighter object (related to $h_2$) during locomotion to ensure safety. As proposed in \cite{lee2023hierarchical}, the safe control command is computed by using the hierarchical safety filter, which is formulated as Quadratic Program (QP) with hierarchical CBFs as follows:
\begin{equation} \label{HCBF-QP}
    \begin{split}
        \bm{u}^{\star} = \argmin_{\bm{u}\in \mathcal{U}} &\quad  \| \bm{u}^{d}  - \bm{u} \|^{2} + W \delta^{2} \\
        \textrm{s.t.} &\quad \dot{h}_1(\bm{x},\bm{u}) \geq - \alpha_1(h_1(\bm{x})), \\
        & \quad \dot{h}_2(\bm{x}, \bm{u}) - \dot{h}_2(\bm{x}, \bm{u}^{i}) = \delta
    \end{split}
\end{equation}
where $W \geq 0$ is a weighting factor in the cost function and $\delta$ accounts for the secondary CBF relaxation. In addition, the intermediate control input $\bm{u}^{i}$ is computed as follows:
\begin{equation}
\begin{split}
    \bm{u}^{i} = \argmin_{\bm{u}\in \mathcal{U}} &\quad \|\bm{u}^{d} - \bm{u}\|^{2} \\
    \textrm{s.t.} &\quad \dot{h}_2(\bm{x}, \bm{u}) \geq - \alpha_2(h_2(\bm{x})).
\end{split}
\end{equation}
The primary safety condition for avoiding the heavier object ($h_1$) is strictly enforced through the inequality $\dot{h}_1(\bm{x}, \bm{u}) \geq -\alpha_1(h_1(\bm{x}))$ in the safety filter \eqref{HCBF-QP}. The violation of the secondary condition ($h_2$), if necessary, is regulated by adjusting the weighting parameter $W$. This parameter is determined through high-fidelity simulations, following a learning-from-demonstration approach, as discussed in \cite{lee2023hierarchical}.

\subsection{Robust H-LIP Control with Disturbance Observer}
\label{section3_3}

In the preliminary section (Section \ref{section2}), we formulated the H-LIP model, which includes a disturbance term $w_{[k]}$ in \eqref{H-LIP}. As discussed, we assume that the disturbance $w_{[k]}$ due to the model discrepancies is negligible when there is no interaction with objects. Thus, we consider that the disturbance $w_{[k]}$ is resulted solely by physical interactions with objects. In this context, when the robot's foot position $u_{[k]}$ significantly deviates from the desired position $u_{[k]}^{d}$ due to the physical interactions, compensating for the disturbance is crucial to ensure robust locomotion and improving the tracking performance. It is important to note that the H-LIP model is linear, and its behaviors in the frontal and sagittal planes are decoupled, making it suitable for implementing a DOB to reduce the effect of physical interactions.

\begin{figure}[t] 
\centering
\includegraphics[width=\linewidth]{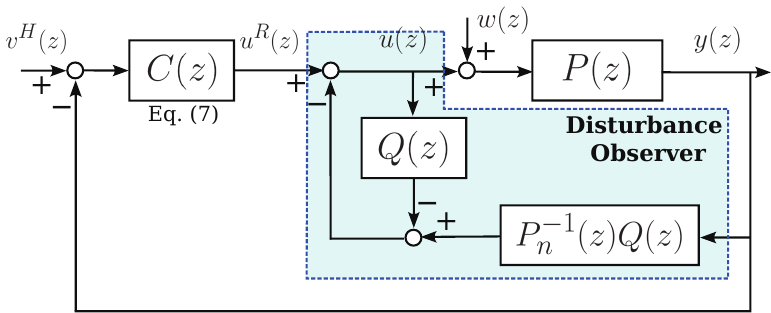}
\vspace{-6mm}
\caption{\textbf{Disturbance Observer}: A typical disturbance observer for a linear system is employed with "Q-filter" since the H-LIP is the nominal system.}
\label{Fig3}
\vspace{-0.6cm}
\end{figure}

Now, we formulate a DOB for the H-LIP system incorporating disturbances as shown in Fig. \ref{Fig3}. First, we consider the CoM velocity ($v_{[k]}^{H}$) as the output to compute the transfer function of our system as follows:
\begin{equation}
    y_{[k+1]}^{R} = \underbrace{\left[ \begin{array}{cc} 0 & 1 \end{array}\right]}_{\mathbf{C}^{H}} \mathbf{x}_{\textrm{SSP}[k+1]}^{R} +  0 \cdot u_{[k]}^{R}. 
\end{equation}
With this input-output relationship, the transfer function of the nominal system is described in $z$ domain.
\begin{equation}
    P_{n}(z) = \mathbf{C}^{H} (z\mathbf{I}  - \mathbf{A}^{H})^{-1}\mathbf{B}^{H}
\end{equation}
In addition, the transfer function of the controller in \eqref{LIP-controller} is denoted as $C(z)$, which is stable since we assume that $\mathbf{A}^{H} + \mathbf{B}^{H}\mathbf{K}_{\textrm{step}}$ is Hurwitz. Let $P(z)$ be the transfer function of an actual linear system. The output can be described as follows in the frequency domain:
\begin{align} \label{output_freq}
    y(z) = & \frac{P_n(z) P(z) C(z)}{G(z)}v^{H}(z) + \frac{P_n(z) P(z) (1-Q(z))}{G(z)} w(z) \nonumber \\
    &  - \frac{P(z)(Q(z)+P_n(z) C(z))}{G(z)}n(z) )
\end{align}
where $G(z) = P_n(z)(1+P(z)C(z)) + Q(z)(P(z)-P_n(z))$. Here, $n(z)$ is the measurement noise, and $Q(z)$ is a stable low-pass filter called "Q-filter". The stability of this filter is ensured by the conditions in \cite{shim2021disturbance}. 

This paper assumes that the measurement noise is negligible, i.e., $n(z) \approx 0$. If $\omega \ll \omega_c$ (such that $Q(z) \approx 1$), the output in \eqref{output_freq} simplifies to: 
\begin{equation}
    y(z) \approx \frac{P_n(z)C(z)}{1+ P_n(z) C(z)} v^{H}(z).
\end{equation}
The control input is computed as follows:
\begin{equation} \label{cmd_dob}
    u_{[k]} = u_{[k]}^{R} + u_{[k-1]}^{Q} - \mathcal{P}_y^{-1}( y_{[k-1]}^{Q})
\end{equation}
where $u_{[k-1]}^{Q}$ and $y_{[k-1]}^{Q}$ are the $k-1$-th input and output filtered by the Q-filter. $\mathcal{P}_{y}$ is a functional mapping from the input $u$ and output $y$ in the time domain. As described in \eqref{cmd_dob}, if $u_{[k-1]}^{Q} - \mathcal{P}_{y}^{-1}(y_{[k-1]}^{Q})$ is negligible, the DOB will not modify the current footstep. However, if there is a significant discrepancy between the filtered previous step and the planned step, the current footstep must be adjusted. It is noted that the adjusted foot step is too large, the robot may lose balance due to the significant interactive force $\bm{f}_{\textrm{int}}$. 
  
\begin{figure*}[t] 
\centering
\includegraphics[width=\linewidth]{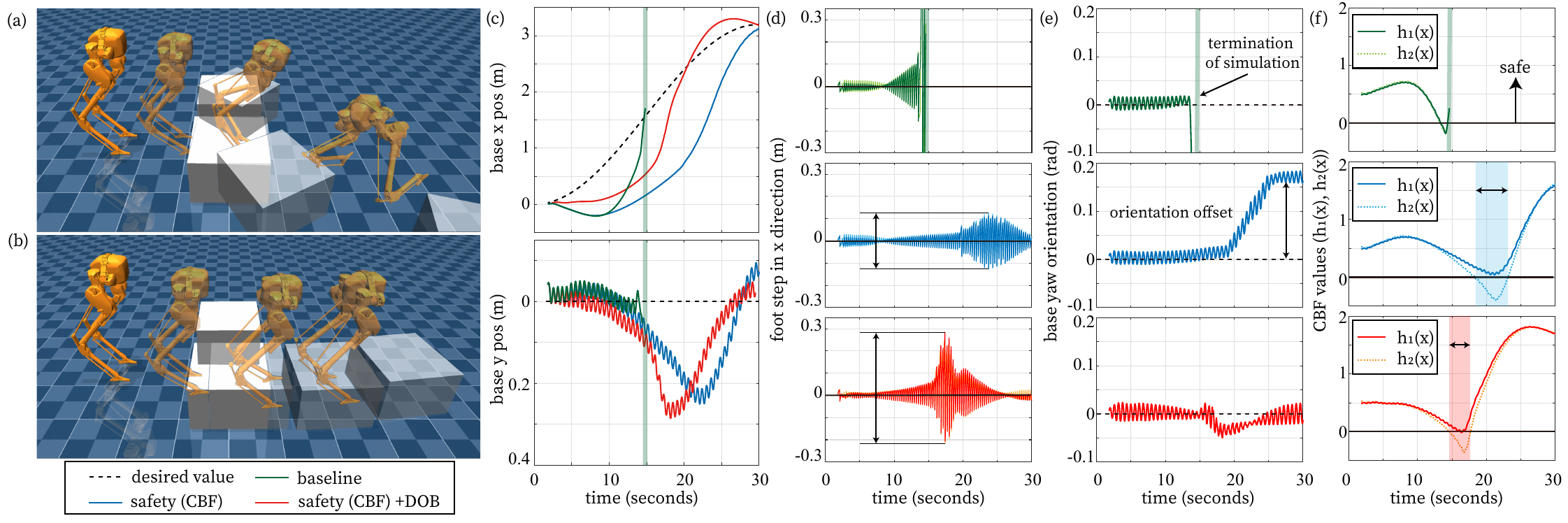}
\vspace{-7mm}
\caption{\textbf{Simulation results}: (a) snapshots of simulation demonstrating baseline locomotion control, (b) snapshot for the proposed locomotion control, (c) x and y positions of the floating base of the robot, (d) footstep of swing foot in the x direction, (e) yaw orientation of the floating base of the robot, (f) CBF values of safety-critical conditions. We implement three locomotion control methods: 1) baseline method (green), 2) CBF-based safe locomotion control (blue), 3) CBF-based safe locomotion control with DOB (red).}
\label{Fig4}
\vspace{-0.6cm}
\end{figure*}

\subsection{Whole-body Controller}
\label{Section3_4}
The low-level controller (WBC) consists primarily of an inverse kinematic controller and dynamic feedforward controller. The full-body dynamics of robots is described by the following equation: 
\begin{equation} \label{full_model}
    \mathbf{D}(\bm{q})\ddot{\bm{q}} + \mathbf{C}(\bm{q}, \dot{\bm{q}}) = \mathbf{S} \bm{\tau} + \mathbf{J}_{c}^{\top}(\mathbf{q}) \bm{f}_{c} 
\end{equation}
where $\mathbf{C}(\bm{q}, \dot{\bm{q}}) \in \mathbb{R}^{n_q}$ represents the sum of Coriolis/centrifugal and gravitational forces and $\mathbf{S} \in \mathbb{R}^{n_q\times (n_q-6)}$ is the selection matrix for the actuated joints. For the inverse kinematics controller, we compute the joint position and velocity command using the projection-based method described in \cite{lee2012intermediate}:
\begin{equation}
\begin{split}
    \bm{q}^{d} =& \bm{q} + \Delta T \left(\mathbf{J}_{\xi|c}(\bm{q}) ^{\dag}k_{\xi}(\bm{\xi}^{d} - \bm{\xi})\right),\\
    \dot{\bm{q}}^{d} =& \mathbf{J}_{\xi|c}(\bm{q})^{\dag} \dot{\bm{\xi}}^{d}
    \end{split}
\end{equation}
where $\mathbf{J}_{\xi|c}(\bm{q}) =  \mathbf{J}_{\xi}(\bm{q}) \mathbf{N}_{c}(\bm{q})$. $\bm{\xi}^{d}$ and $\dot{\bm{\xi}}^{d}$ are the desired output specification and its time-derivative, respectively. $\Delta T$ and $k_{\xi}$ are the control loop time interval and the proportional gain for reducing the output error. During the single support phase, the output is defined as follows:
\begin{equation}
    \begin{split}
        \bm{\xi} = \textrm{Vertcat}(& \bm{\xi}_{\textrm{com}}^{\textrm{pos}}, \bm{\xi}_{\textrm{st/hip,z}}^{\textrm{ori}}, \bm{\xi}_{\textrm{st/foot,y}}^{\textrm{ori}}, \bm{\xi}_{\textrm{pelvis,xy}}^{\textrm{ori}}, \\
        & \bm{\xi}_{\textrm{foot}}^{\textrm{pos}}, \bm{\xi}_{\textrm{sw/hip,z}}^{\textrm{ori}}, \bm{\xi}_{\textrm{sw/foot,y}}^{\textrm{ori}}) \in \mathbb{R}^{12}
    \end{split}
\end{equation}
where $\textrm{Vertcat}(.)$ denotes the vertically concatenated vector of input vectors. $\bm{\xi}_{\textrm{com}}$, $\bm{\xi}_{\textrm{sw/foot}}$, $\bm{\xi}_{\textrm{st/foot}}$, $\bm{\xi}_{\textrm{sw/hip}}$, $\bm{\xi}_{\textrm{pelvis}}$, and $\bm{\xi}_{\textrm{st/hip}}$ denote the outputs related to CoM, swing foot, standing foot, hip of swing leg, pelvis, and hip of standing leg, respectively. The superscripts $(.)^{\textrm{pos}}$ and $(.)^{\textrm{ori}}$ represent the position and orientation parts of the output $(.)$. For preventing the conflicts between the outputs, we find the rank of $\mathbf{J}_{\xi|c}(\bm{q})$ and compact singular value decomposition is applied if the matrix becomes singular. 

The kinematic feedback controller is computed as a PD controller as follows:
\begin{equation}
    \bm{\tau}_{\textrm{kin}} = \mathbf{K}_{p}(\bm{q}^{d} - \bm{q}) + \mathbf{K}_{d}(\dot{\bm{q}}^{d} - \dot{\bm{q}})
\end{equation}
where $\mathbf{K}_p$ and $\mathbf{K}_d$ are the proportional and derivative gains, respectively. Dynamic whole-body controllers typically include complex dynamic constraints based on the equation of motion in \eqref{full_model} to obtain the feedforward torque control input. During dynamic walking, the feedforward control input $\hat{\bm{\tau}}$ is computed to compensate for the gravitational force, assisted by a kinematic feedback controller for fast computation, ensuring minimal latency in the control loop. The final control command torque is given by $\bm{\tau}_{\textrm{cmd}} = \bm{\tau}_{\textrm{kin}} + \hat{\bm{\tau}}$. 

\section{Implementation}
\label{section4}
We demonstrate high-fidelity simulations with a biped robot ``Cassie'' to validate the proposed framework. The simulation scenario involves navigating toward a predefined destination, while passing through a region obstructed by two boxes of identical size but different weights-$15$ kg and $5$ kg.

\subsection{Simulations}
To validate our proposed approach, we utilized mujoco physics engine \cite{todorov2012mujoco} to implement high-fidelity dynamic simulations. For a fair evaluation of the proposed method, we also implement a baseline locomotion method consisting of walking pattern generation based on the H-LIP model and WBC, excluding mass estimation, safety-critical planning, and DOB. For the H-LIP model, the desired velocity reference is computed using a simple P control law: $v^{R,d} = K_{p,\textrm{H-LIP}}(p^{R,d} - p^{R})$ where $K_{p,\textrm{H-LIP}}$ is the proportional gain for generating the reference velocity. The destination is $[3.2, \; 0]$ m, and a reference trajectory of the pelvis position is generated using cubic spline interpolation. Two boxes, each with dimensions $L\times W \times H =$ 0.3 m $\times$ 0.3 m $\times$ 0.3 m are placed at $[1,\; -0.4]$ m and $[1,\; 0.4]$ m along the path. 

In the simulation results, the baseline method leads to the robot's failure, as it falls after colliding with the heavier box, as shown in Fig. \ref{Fig4}(a). In contrast, the proposed layered framework, both with and without DOB, allows the robot to avoid the heavier box and successfully reach the goal by kicking the lighter one aside, as depicted in Fig. \ref{Fig4}(b).

\begin{figure}[t] 
\centering
\includegraphics[width=0.9\linewidth]{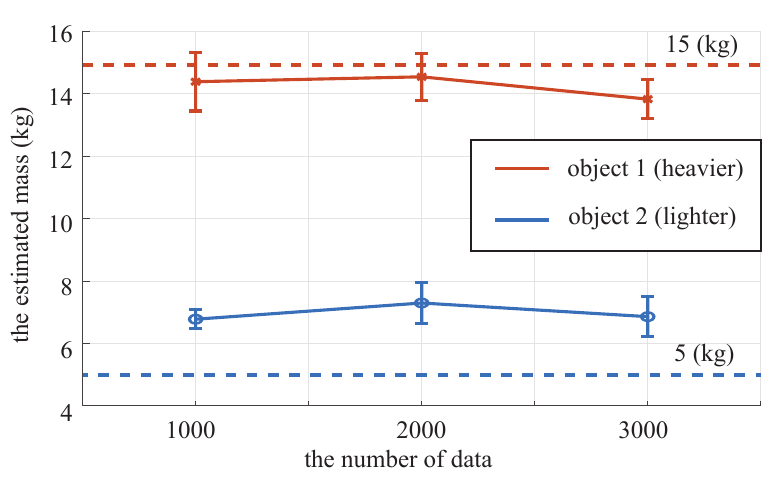}
\vspace{-4mm}
\caption{\textbf{Approximated estimation of mass}: Three data sets (1000, 2000, 3000) are utilized to estimate the mass of each obstacle. }
\label{Fig5}
\vspace{-0.6cm}
\end{figure}

More specifically, we analyze the robot’s performance operated by the proposed framework with and without DOB. As illustrated in Fig. \ref{Fig4}(c), both safety-critical locomotion control methods, with and without DOB, succeed in navigating to the goal by overcoming the obstacles. However, the DOB-enabled method shows a significantly lower tracking error in base yaw orientation ($0.01$ rad) compared to the method without DOB ($0.17$ rad). This is because DOB generates larger footstep lengths (maximum of $0.295$ m) to compensate for disturbances, compared to the method without DOB (maximum of $0.124$ m), as shown in Fig. \ref{Fig4}(d).

\subsection{Discussion}
In these high-fidelity simulations, we estimate the mass of the two objects using three data sets, as shown in Fig. \ref{Fig5}. To improve estimation accuracy, we discarded the peak values generated at the moment of impact. Although the estimated masses are different from the actual values, they are sufficient to determine which object is heavier, allowing us to use the approximated values to establish the hierarchy of CBFs. However, to improve the sensitivity and accuracy of mass estimation in real hardware experiments, we plan to incorporate additional sensors and perceptual systems. Moreover, accounting for rotational behavior and momentum could enhance the precision of the estimation.

Regarding safety, as shown in Fig. \ref{Fig4}(f), the CBF-based safety filter effectively enforces the most critical safety condition ($h_1(x) > 0$), even if the secondary condition is temporarily violated ($h_2(x) < 0$). In contrast, the baseline method fails to reach the destination due to physical interactions with both objects, leading to violations of both safety conditions ($h_1(x) < 0$ and $h_2(x) < 0$ in the first plot of Fig. \ref{Fig4}(f)). In addition, we verify that the DOB with H-LIP reduces the tracking error when disturbances occur due to unavoidable interactions with the environment.

\section{Conclusion}
This paper proposes a multi-layered framework for the safety-critical locomotion of biped robots in unstructured environments where obstacles must be removed to reach a goal. In the top layer, we leverage an estimation process to compare the dynamic properties of the objects. Based on the estimation results, a global path is refined that prioritizes avoiding heavier object by imposing a hierarchy of safety-critical conditions while navigating toward the goal. To enhance the stability during locomotion, a DOB compensates for the disturbances caused by the interaction with obstacle. This approach generates more stable and robust bipedal locomotion compared to the baseline method, as demonstrated through simulations.

Future work will focus on advancing the framework by incorporating perception for navigating more complex environments with multiple unknown obstacles. Additionally, various learning techniques will be explored to better interpret unstructured environments. The proposed layered architecture will also be extended to enable humanoid robots to autonomously perform locomanipulation tasks.

\bibliographystyle{IEEEtran}
\balance
\bibliography{l_css}

\end{document}